\documentclass[11pt,letterpaper]{article}
\usepackage{naaclhlt2015}

\usepackage{times}
\usepackage{latexsym}
\usepackage{amsmath}
\usepackage{multirow}
\usepackage{url}
\usepackage{amsfonts}
\usepackage{amssymb}
\usepackage{graphicx}
\usepackage{mathrsfs}
\usepackage{tablefootnote}
\usepackage{arydshln}

\title{Compositional Distributional Semantics with Long Short Term Memory}

\author{Phong Le \and Willem Zuidema\\
Institute for Logic, Language and Computation \\
University of Amsterdam, the Netherlands\\
{\tt \{p.le,zuidema\}@uva.nl } \\ }

\begin{document}

\maketitle

\begin{abstract}
	
	We are proposing an extension of the recursive neural network that makes 
	use of a variant of the long short-term memory architecture. The extension 
	allows information low in parse trees to be stored in a memory register 
	(the `memory cell') and used much later higher up in the parse tree. 
	This provides a solution to the vanishing gradient problem and allows 
	the network to capture long range dependencies. Experimental results show 
	that our composition outperformed the traditional neural-network 
	composition on the Stanford Sentiment Treebank. 

\end{abstract}

\section{Introduction}

Moving from lexical to compositional semantics in vector-based semantics requires
answers to two difficult questions: (i) what is the nature of the composition 
functions (given that the lambda calculus for variable binding is no longer applicable), 
and (ii) how do we learn the parameters of those functions (if they have any) from data? 
A number of classes of functions have been proposed in answer to the first question, 
including simple linear functions like vector addition \cite{mitchell2009language},
non-linear functions like those defined by multi-layer neural networks 
\cite{socher_learning_2010}, and vector matrix multiplication and tensor linear 
mapping \cite{baroni_frege_2013}. The matrix and tensor-based functions have the 
advantage of allowing a relatively straightforward comparison with formal semantics, 
but the fact that multi-layer neural networks with non-linear activation functions 
like sigmoid can approximate any continuous function \cite{cybenko1989approximation}
already make them an attractive choice.

In trying to answer the second question, the advantages of approaches based on 
neural network architectures, such as the recursive neural network (RNN) model \cite{socher2013recursive} 
and the convolutional neural network model \cite{kalchbrenner-grefenstette-blunsom:2014:P14-1}, 
are even clearer. Models in this paradigm can take advantage of general learning 
procedures based on back-propagation, and with the rise of `deep learning', 
of a variety of efficient algorithms and tricks to further improve training. 

Since the first success of the RNN model \cite{socher_parsing_2011} in constituent parsing, 
two classes of extensions have been proposed. 
One class is to enhance its compositionality by using tensor product
\cite{socher2013recursive} or concatenating RNNs horizontally to make a deeper net
\cite{irsoy2014deep}. The other is to extend its topology in order to fulfill a wider 
range of tasks, like \newcite{le2014the} for dependency parsing and 
\newcite{paulus2014global} for context-dependence sentiment analysis. 

Our proposal in this paper is an extension of the RNN model to improve compositionality. 
Our motivation is that, like training recurrent neural networks, 
training RNNs on deep trees can suffer from the vanishing gradient problem
\cite{Hochreiter2001}, 
i.e., that errors propagated back to the leaf nodes shrink exponentially. 
In addition, information sent from a leaf node to the root can be obscured 
if the path between them is long, thus leading to the problem how to capture 
long range dependencies. We therefore borrow the long short-term memory (LSTM) architecture 
\cite{hochreiter1997long} from recurrent neural network research to tackle those two problems.
The main idea is to allow  information low in a parse tree to be stored in 
a memory cell and used much later higher up in the parse tree,
by recursively \textit{adding} up all memory 
into memory cells in a bottom-up manner. In this way, errors propagated 
back through structure do not vanish. And information from leaf nodes
is still (loosely) preserved and can be used directly at any higher nodes in 
the hierarchy. 
We then apply this composition to sentiment analysis. Experimental 
results show that the new composition works better than the traditional 
neural-network-based composition.

The outline of the rest of the paper is as follows. We first, in Section~\ref{section background}, give a brief 
background on neural networks, including the multi-layer neural network, 
recursive neural network, recurrent neural network, and 
LSTM. 
We then propose the LSTM for recursive neural networks in Section~\ref{section lstm}, 
and its application to sentiment analysis in Section~\ref{section lstm sent}. 
Section~\ref{section experiments} shows our experiments.

\section{Background}
\label{section background}

\subsection{Multi-layer Neural Network}

\begin{figure}[t]
    \centering
    \includegraphics[width=0.45\textwidth]{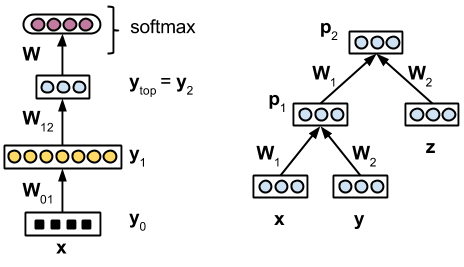}
    \caption{Multi-layer neural network (left) and Recursive neural network (right).
    Bias vectors are removed for the simplicity. }
    \label{figure net}
\end{figure}

\begin{figure}[t]
    \centering
    \includegraphics[width=0.45\textwidth]{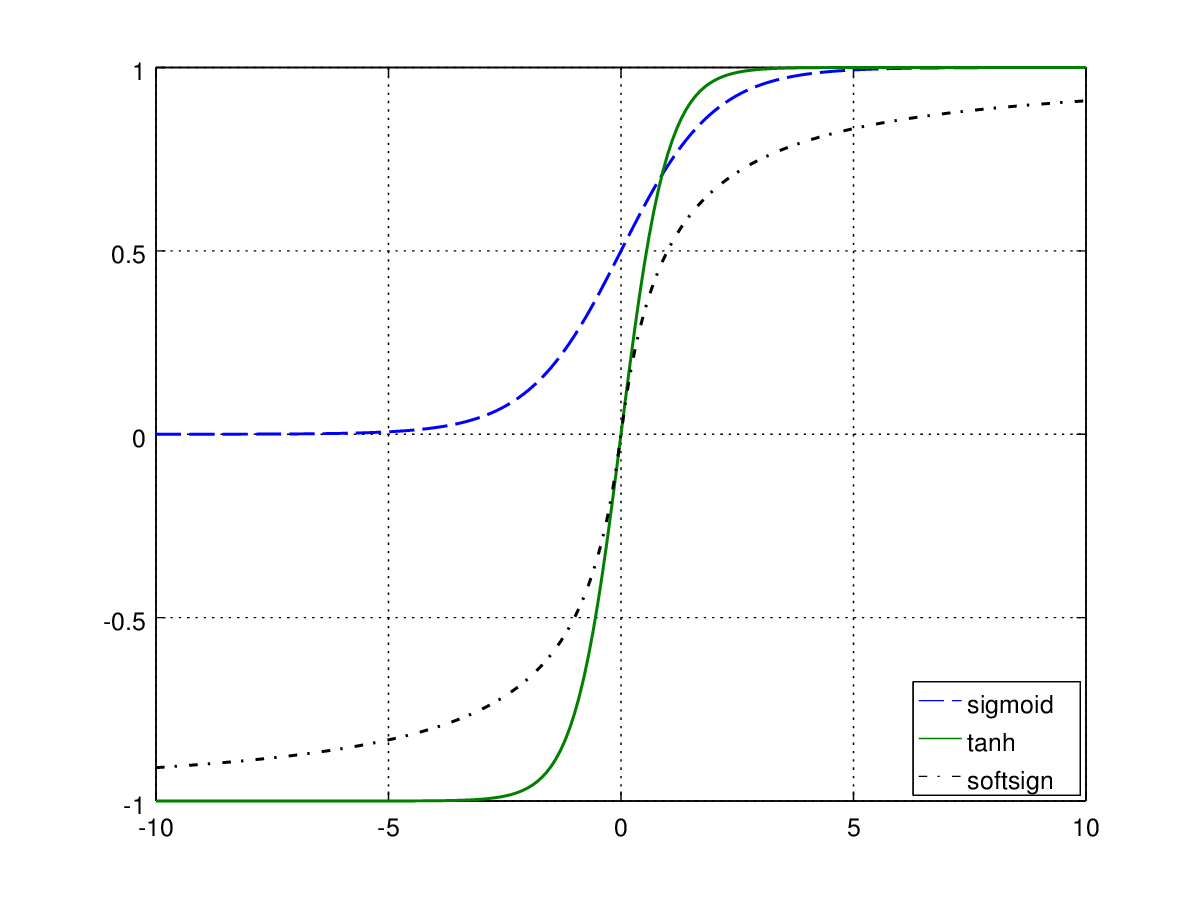}
    \caption{Activation functions: $\text{sigmoid}(x) = \frac{1}{1+e^{-x}}$, $\tanh(x) = \frac{e^{2x}-1}{e^{2x}+1}$, 
    $\text{softsign}(x) = \frac{x}{1 + |x|}$.}
    \label{figure act-func}
\end{figure}

In a multi-layer neural network (MLN), neurons are organized in layers 
(see Figure~\ref{figure net}-left). A neuron in layer $i$ receives signal from neurons 
in layer $i-1$ and transmits its output to neurons in layer $i+1$. 
\footnote{This is a simplified definition. In practice, any layer $j < i$ can connect 
to layer $i$.}
The computation is given by
\begin{equation*}
    \mathbf{y}_i = g \big( \mathbf{W}_{i-1,i} \mathbf{y}_{i-1} + \mathbf{b}_i \big)
\end{equation*}
where real vector $\mathbf{y}_i$ contains the activations of the neurons in layer $i$;
$\mathbf{W}_{i-1,i} \in \mathbb{R}^{|\mathbf{y}_i| \times |\mathbf{y}_{i-1}|}$ is the matrix of weights of 
connections from  
layer $i-1$ to layer $i$; $\mathbf{b}_i \in \mathbb{R}^{|\mathbf{y}_i|}$ is the vector of biases 
of the neurons in layer $i$; $g$ is an activation function, e.g. \textit{sigmoid}, 
\textit{tanh}, or \textit{softsign} (see Figure~\ref{figure act-func}). 

For classification tasks, we put a \textit{softmax} layer on the top of the 
network, and compute the probability of assigning a class $c$ to an input
$\mathbf{x}$ by 
\begin{equation}
    Pr(c | \mathbf{x}) = \text{softmax}(c) = \frac{e^{u(c,\mathbf{y}_{\text{top}})}}{\sum_{c' \in C} e^{u(c',\mathbf{y}_{\text{top}})}}
    \label{equation softmax}
\end{equation}
where 
$\left[ u(c_1,\mathbf{y}_{\textit{top}}),...,u(c_{|C|},\mathbf{y}_{\textit{top}}) \right]^T = \mathbf{W} \mathbf{y}_{\textit{top}} + \mathbf{b}$; $C$ is the 
set of all possible classes; 
$\mathbf{W} \in \mathbb{R}^{|C| \times |\mathbf{y}_{top}|}, \mathbf{b}\in \mathbb{R}^{|C|}$ 
are a weight matrix and a bias vector.

Training an MLN is to minimize an objective function $J(\theta)$
where $\theta$ is the parameter set (for classification, $J(\theta)$
is often a negative log likelihood). 
Thanks to the back-propagation algorithm \cite{rumelhart1988learning}, the gradient 
$\partial J / \partial \theta$ is efficiently computed; 
the gradient descent method thus is used to minimize $J$.

\subsection{Recursive Neural Network}
\label{subsection rnn}

A recursive neural network (RNN) \cite{goller_learning_1996} is an MLN where, given 
a tree structure, we recursively apply the same weight matrices at each inner node 
in a bottom-up manner. In order to see how an RNN works,  
consider the following example. Assume that there is a constituent with parse
tree $(p_2 \; (p_1 \; x \; y) \; z)$ (Figure~\ref{figure net}-right), and 
that $\mathbf{x},\mathbf{y},\mathbf{z} \in \mathbb{R}^{d}$ are
the vectorial representations of the three words $x$, $y$ and $z$, respectively. 
We use a neural network which consists of a weight matrix 
$\mathbf{W}_1  \in \mathbb{R}^{d \times d}$ for left children and 
a weight matrix $\mathbf{W}_2  \in \mathbb{R}^{d \times d}$ for 
right children to compute the vector for a parent node in a bottom up manner. 
Thus, we  compute $p_1$
\begin{equation}
	\mathbf{p}_1 = g(\mathbf{W}_1 \mathbf{x} + \mathbf{W}_2 \mathbf{y} + \mathbf{b})
	\label{equation rnn p1}
\end{equation}
where $\mathbf{b}$ is a bias vector and $g$ is an activation function.
Having computed $p_1$, we can then move one level up in the hierarchy and compute $p_2$:
\begin{equation}
	\mathbf{p}_2 = g(\mathbf{W}_1 \mathbf{p}_1 + \mathbf{W}_2 \mathbf{z} + \mathbf{b})
	\label{equation rnn p2}
\end{equation}
This process is continued until we reach the root node.

Like training an MLN, training an RNN uses the gradient descent method
to minimize an objective function $J(\theta)$. The gradient 
$\partial J / \partial \theta$ is efficiently computed thanks to the back-propagation 
through structure algorithm \cite{goller_learning_1996}.

The RNN model and its extensions have been 
employed successfully to solve a wide range of problems: from parsing (constituent parsing 
\cite{socher2013parsing}, dependency parsing \cite{le2014the}) to classification 
(e.g. sentiment analysis \cite{socher2013recursive,irsoy2014deep}, 
paraphrase detection \cite{socher_dynamic_2011}, semantic role labelling \cite{leinside}).

\subsection{Recurrent Networks and Long Short-Term Memory}

\begin{figure}
    \centering
    \includegraphics[width=0.45\textwidth]{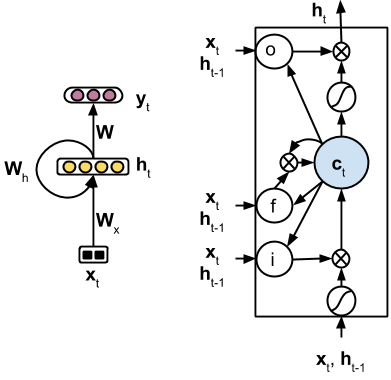}
    \caption{Simple recurrent neural network (left) and long short-term memory (right).
    Bias vectors are removed for the simplicity. }
    \label{figure rrnn}
\end{figure}

A neural network is \emph{recurrent} if it has at least one directed ring 
in its structure. 
In the natural language processing field, the simple recurrent neural network (SRN) proposed by 
\newcite{elman_finding_1990} (see Figure~\ref{figure rrnn}-left) 
and its extensions are used to tackle sequence-related problems, such as 
machine translation \cite{sutskever2014sequence} and language modelling \cite{mikolov2010recurrent}. 

In an SRN, an input $\mathbf{x}_t$ is fed to the network at each time $t$. 
The hidden layer $\mathbf{h}$, which has activation $\mathbf{h}_{t-1}$ right before  
$\mathbf{x}_t$ comes in, plays a role as a memory store capturing the whole history 
$\big(\mathbf{x}_0,...,\mathbf{x}_{t-1}\big)$. When $\mathbf{x}_t$ comes in, 
the hidden layer updates its activation by
\begin{equation*}
    \mathbf{h}_t = g\big(\mathbf{W}_x \mathbf{x}_t + \mathbf{W}_h \mathbf{h}_{t-1} + 
                    \mathbf{b} \big)
\end{equation*}
where $\mathbf{W}_x \in \mathbb{R}^{|\mathbf{h}| \times |\mathbf{x}_t|}$, $\mathbf{W}_h \in \mathbb{R}^{|\mathbf{h}| \times |\mathbf{h}|}$, $\mathbf{b} \in \mathbb{R}^{|\mathbf{h}|}$ are weight matrices and a bias vector; 
$g$ is an activation.

This network model thus, in theory, can 
be used to estimate probabilities conditioning on long histories. 
And computing gradients is efficient thanks to the back-propagation 
through time algorithm \cite{werbos1990backpropagation}. 
In practice, however, training recurrent neural networks with the gradient descent method is 
challenging because gradients $\partial J_t / \partial \mathbf{h}_j$ 
($j \leq t$, $J_t$ is the objective function at time $t$)
vanish quickly after a few back-propagation steps \cite{Hochreiter2001}. 
In addition, it is difficult to capture long range dependencies, i.e. the output 
at time $t$ depends on some inputs that happened very long time ago.
One solution for this, proposed by \newcite{hochreiter1997long} and enhanced by 
\newcite{gers2001long}, 
is \textit{long short-term memory} (LSTM).

\paragraph{Long Short-Term Memory} 
The main idea of the LSTM architecture 
is to maintain a memory of \emph{all} inputs the hidden layer received over time, by \emph{adding up}
all (gated) inputs to the hidden layer through time 
to a memory cell. In this way, errors propagated back through time 
do not vanish and even inputs received a very long time ago 
are still (approximately) preserved and can play a role in computing the output of the network
(see the illustration in \newcite[Chapter 4]{graves2012supervised}).

An LSTM cell (see Figure~\ref{figure rrnn}-right) consists of a memory cell 
$c$, an input gate $i$, a forget gate $f$, an output gate $o$. 
Computations occur in this cell are given below
\begin{align*}
    \mathbf{i}_t &= \sigma \big(\mathbf{W}_{xi} \mathbf{x}_t + \mathbf{W}_{hi} \mathbf{h}_{t-1}
                    + \mathbf{W}_{ci} \mathbf{c}_{t-1} + \mathbf{b}_i \big) \\
    \mathbf{f}_t &= \sigma \big(\mathbf{W}_{xf} \mathbf{x}_t + \mathbf{W}_{hf} \mathbf{h}_{t-1}
                    + \mathbf{W}_{cf} \mathbf{c}_{t-1} + \mathbf{b}_f \big) \\
    \mathbf{c}_t &= \mathbf{f}_t \odot \mathbf{c}_{t-1} + \\
    			& \;\;\;\;\; \mathbf{i}_t \odot \tanh \big(\mathbf{W}_{xc} \mathbf{x}_t + \mathbf{W}_{hc} \mathbf{h}_{t-1} + \mathbf{b}_c \big) \\
    \mathbf{o}_t &= \sigma \big(\mathbf{W}_{xo} \mathbf{x}_t + \mathbf{W}_{ho} \mathbf{h}_{t-1}
                    + \mathbf{W}_{co} \mathbf{c}_{t} + \mathbf{b}_o \big) \\
    \mathbf{h}_t &= \mathbf{o}_t \odot \tanh(\mathbf{c}_t)
\end{align*}
where $\sigma$ is the sigmoid function; $\mathbf{i}_t$, $\mathbf{f}_t$, $\mathbf{o}_t$
are the outputs (i.e. activations) of the corresponding gates; $\mathbf{c}_t$ is the state
of the memory cell; $\odot$ denotes the element-wise multiplication operator; 
$\mathbf{W}$'s and $\mathbf{b}$'s are weight matrices and bias vectors.

Because the sigmoid function has the output range $(0,1)$ (see Figure~\ref{figure act-func}), 
activations of those gates can be seen as normalized weights. Therefore, 
intuitively, the network can learn to use the input gate to decide when to memorize information, 
and similarly learn to use the output gate to decide
when to access that memory. The forget gate, finally, is to reset the memory.

\section{Long Short-Term Memory in RNNs}
\label{section lstm}

\begin{figure}
    \centering
    \includegraphics[width=0.45\textwidth]{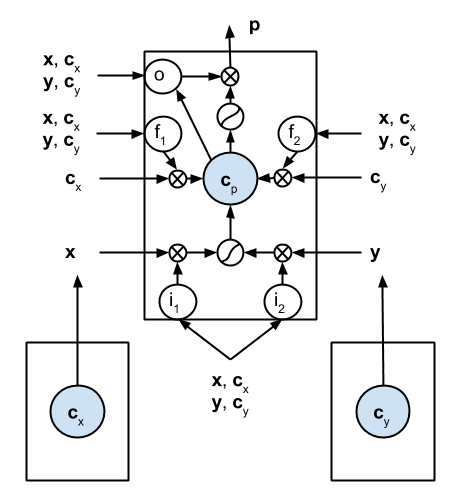}
    \caption{Long short-term memory for recursive neural network.}
    \label{figure lstm-rnn}
\end{figure}

In this section, we propose an extension of the LSTM for the RNN model
(see Figure~\ref{figure lstm-rnn}). A key feature of the RNN is to hierarchically 
combine information from two children to compute the parent vector; the idea in this section 
is to extend the LSTM such that not only the output from each of the children is used, 
but also the contents of their memory cells. This way, the network has the option 
to store information when processing constituents low in the parse tree, 
and make it available later on when it is processing constituents high in the parse tree.

For the simplicity 
\footnote{Extending our LSTM for $n$-ary trees is trivial. }, 
we assume that the parent node $p$ has two children $a$
and $b$. The LSTM at $p$ thus has two input gates $i_1,i_2$ and two forget gates
$f_1,f_2$ for the two children. Computations occuring in this LSTM are:

\begin{align*}
\mathbf{i}_1 &= \sigma \big(\mathbf{W}_{i1} \mathbf{x} + \mathbf{W}_{i2} \mathbf{y}
                + \mathbf{W}_{ci1} \mathbf{c}_x + \mathbf{W}_{ci2} \mathbf{c}_y + \mathbf{b}_i \big) \\
\mathbf{i}_2 &= \sigma \big(\mathbf{W}_{i1} \mathbf{y} + \mathbf{W}_{i2} \mathbf{x}
                + \mathbf{W}_{ci1} \mathbf{c}_y + \mathbf{W}_{ci2} \mathbf{c}_x + \mathbf{b}_i \big) \\
\mathbf{f}_1 &= \sigma \big(\mathbf{W}_{f1} \mathbf{x} + \mathbf{W}_{f2} \mathbf{y}
                + \mathbf{W}_{cf1} \mathbf{c}_x + \mathbf{W}_{cf2} \mathbf{c}_y + \mathbf{b}_f \big) \\
\mathbf{f}_2 &= \sigma \big(\mathbf{W}_{f1} \mathbf{y} + \mathbf{W}_{f2} \mathbf{x}
                + \mathbf{W}_{cf1} \mathbf{c}_y + \mathbf{W}_{cf2} \mathbf{c}_x + \mathbf{b}_f \big) \\
\mathbf{c}_p &= \mathbf{f}_1 \odot \mathbf{c}_x + \mathbf{f}_2 \odot \mathbf{c}_y + \\
                & \;\;\;\;\; g \big(\mathbf{W}_{c1} \mathbf{x} \odot \mathbf{i}_1 + 
                            \mathbf{W}_{c2} \mathbf{y} \odot \mathbf{i}_2 + \mathbf{b}_c \big) \\
\mathbf{o} &= \sigma \big(\mathbf{W}_{o1} \mathbf{x} + \mathbf{W}_{o2} \mathbf{y}
                + \mathbf{W}_{co} \mathbf{c} + \mathbf{b}_o \big) \\
\mathbf{p} &= \mathbf{o} \odot g(\mathbf{c}_p)
\end{align*}
where $\mathbf{u}$ and $\mathbf{c}_u$ are the output and the state of the memory cell at 
node $u$; $\mathbf{i}_1$, $\mathbf{i}_2$, $\mathbf{f}_1$, $\mathbf{f}_2$, $\mathbf{o}$ 
are the activations of the corresponding gates; $\mathbf{W}$'s and $\mathbf{b}$'s are 
weight matrices and bias vectors; and $g$ is an activation function.

Intuitively, the input gate $i_j$ lets the LSTM at the parent node decide how important the output 
at the $j$-th child is. If it is important, the input gate $i_j$ will have an activation close to $1$. 
Moreover, the LSTM controls, using the forget gate $f_j$, the degree to which information from the memory of the $j$-th child should be 
added to its memory. 

Using one input gate and one forget gate for each child makes the LSTM
flexible in storing memory and computing composition. For instance, 
in a complex sentence containing a main clause and a dependent clause it could be beneficial 
if only information about the main clause is passed on to higher levels. This can be achieved 
by having low values for the input gate and the forget gate for the child node that covers 
the dependent clause, and high values for the gates corresponding to the child node covering 
(a part of) the main clause. More interestingly, this LSTM can even allow a child to contribute to
composition by activating the corresponding input gate, but ignore the child's memory 
by deactivating the corresponding forget gate. This happens when the information given by the child is
temporarily important only.

\section{LSTM-RNN model for Sentiment Analysis
\footnote{The LSTM architecture was already applied to the sentiment analysis task, 
for instance in the model proposed at \url{http://deeplearning.net/tutorial/lstm.html}.
Independently from and concurrently with our work, \newcite{tai2015improved} and 
\newcite{zhu2015long} have developed very similar models applying LTSM to RNNs.}}
\label{section lstm sent}

\begin{figure*}[t]
    \centering
    \includegraphics[width=0.8\textwidth]{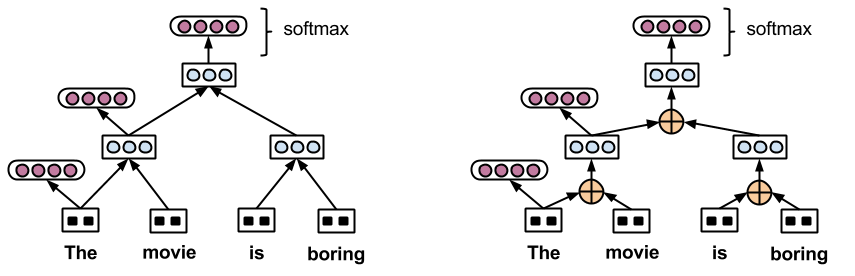}
    \caption{The RNN model (left) and LSTM-RNN model (right) for sentiment analysis.}
    \label{figure sentiment-lstm-rnn}
\end{figure*}

In this section, we introduce a model using the proposed LSTM for 
sentiment analysis. 
Our model, named LSTM-RNN, is an extension of the traditional RNN model 
(see Section~\ref{subsection rnn}) 
where traditional composition function $g$'s in 
Equations~\ref{equation rnn p1}-~\ref{equation rnn p2}
are replaced by our proposed LSTM (see Figure~\ref{figure sentiment-lstm-rnn}).
On top of the node covering a phrase/word, if its sentiment class 
(e.g. positive, negative, or neutral) is available, we put a softmax layer 
(see Equation~\ref{equation softmax}) to compute the probability of assigning a class to it.

The vector representations of words (i.e. word embeddings) can be 
initialized randomly, or pre-trained. 
The memory of any leaf node $w$, i.e. $\mathbf{c}_w$, is 0.

Similarly to \newcite{irsoy2014deep}, we `untie' leaf nodes and inner nodes: we use one weight matrix set
for leaf nodes and another set for inner nodes. Hence, let $d_w$ and $d$ respectively 
be the dimensions of word embeddings (leaf nodes) and vector representations of 
phrases (inner nodes), 
all weight matrices from a leaf node to an inner node have size $d \times d_w$, 
and all weight matrices from an inner node to another inner node have size $d \times d$.

\paragraph{Training} Training this model is to minimize the following objective function, which is the 
cross-entropy over training sentence set $\mathcal{D}$ plus an L2-norm regularization term
\begin{equation*}
J(\theta) = - \frac{1}{|\mathcal{D}|} \sum_{s \in \mathcal{D}} \sum_{p \in s} \log Pr(c_p | \mathbf{p} ) + \frac{\lambda}{2} || \theta ||^2 
\end{equation*}
where $\theta$ is the parameter set, $c_p$ is the sentiment class of phrase $p$, 
$\mathbf{p}$ is the vector representation at the node covering $p$, 
$Pr(c_p | \mathbf{p})$ is computed by the softmax function, and $\lambda$ is the regularization 
parameter. Like training an RNN, we use the mini-batch gradient descent method to minimize 
$J$, where the gradient $\partial J / \partial \theta$ is computed efficiently 
thanks to the back-propagation through structure \cite{goller_learning_1996}. We use the AdaGrad method \cite{duchi2011adaptive} 
to automatically update the learning rate for each parameter. 

\subsection{Complexity}
We analyse the complexities of the RNN and LSTM-RNN models in the forward phase, 
i.e. computing vector representations for inner nodes and classification probabilities.
The complexities in the backward phase, i.e. computing gradients 
$\partial J / \partial \theta$, can be analysed similarly. 

The complexities of the two models are dominated by the matrix-vector multiplications 
that are carried out. Since the number of sentiment classes is very small 
(5 or 2 in our experiments) compared to $d$ and $d_w$, 
we only consider those matrix-vector multiplications 
which are for computing vector representations at the inner nodes. 

For a sentence consisting of $N$ words, assuming that its parse tree is
binarized without any unary branch (as in the data set we use in our experiments), 
there are $N-1$ inner nodes, $N$ links from leaf nodes to inner nodes, and $N-2$ links
from inner nodes to other inner nodes. The complexity of RNN in the forward 
phase is thus approximately 
\begin{equation*}
    N \times d \times d_w + (N-2) \times d \times d 
\end{equation*}
The complexity of LSTM-RNN is approximately 
\begin{equation*}
    N \times 6 \times d \times d_w   + (N-2) \times 10 \times d \times d + (N-1) \times d \times d
\end{equation*}
If $d_w \approx d$, the complexity of LSTM-RNN is about 8.5 times higher than 
the complexity of RNN.

In our experiments, this difference is not a problem because training and evaluating the 
LSTM-RNN model is very fast: it took us, on a single core
of a modern computer, about 10 minutes to train the model ($d=50, d_w = 100$) on 
8544 sentences, and about 2 seconds to evaluate it on 2210 sentences.

\section{Experiments}
\label{section experiments}

\subsection{Dataset}

We used the Stanford Sentiment Treebank\footnote{\url{http://nlp.stanford.edu/sentiment/treebank.html}}
\cite{socher2013recursive} which consists of 5-way fine-grained sentiment labels 
(very negative, negative, neutral, positive, very positive) for 215,154 phrases of 
11,855 sentences. The standard splitting is also given: 
8544 sentences for training, 1101 for development, and 2210 for testing.
The average sentence length is 19.1. 

In addition, the treebank also supports binary sentiment (positive, negative) 
classification by removing neutral labels, leading 
to: 6920 sentences for training, 872 for development, and 1821 for testing.

The evaluation metric is the accuracy, given by $\frac{100 \times \#correct}{ \#total}$.

\subsection{LSTM-RNN vs. RNN}

\paragraph{Setting}
We initialized the word vectors by the 100-D GloVe\footnote{\url{http://nlp.stanford.edu/projects/GloVe/}} 
word embeddings \cite{pennington2014GloVe}, which were trained on a 6B-word corpus. 
The initial values for a weight matrix were uniformly sampled from the symmetric interval
$\big[{-\frac{1}{\sqrt{n}}} , \frac{1}{\sqrt{n}} \big]$ where $n$ is the number of total input units.

For each model (RNN and LSTM-RNN), we tested three activation functions: softmax, tanh, 
and softsign, leading to six sub-models.
Tuning those sub-models on the development set, we chose the dimensions of vector representations 
at inner nodes $d = 50$, learning rate $0.05$, regularization parameter $\lambda = 10^{-3}$, 
and mini-batch-size 5. 

On each task, we run each sub-model 10 times. 
Each time, we trained the sub-model in 20 epochs and selected the network 
achieving the highest accuracy on the development set.

\begin{figure}[t]
\centering
\includegraphics[width=0.5\textwidth]{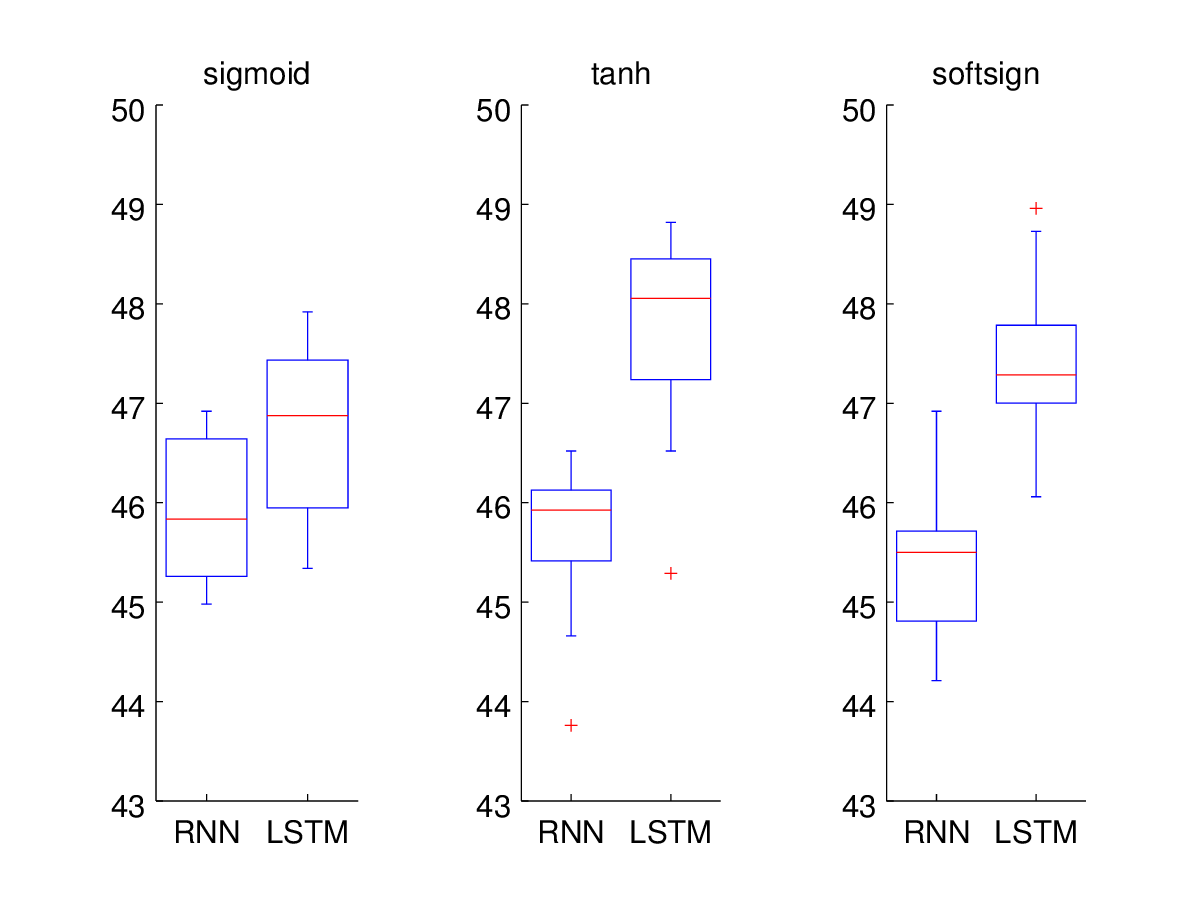}
\caption{Boxplots of accuracies of 10 runs of RNN and LSTM-RNN on the test set in the fine-grained classification task.
(LSTM stands for LSTM-RNN.) }
\label{figure 5c LSTM vs RNN}
\end{figure}

\begin{figure}[t]
\centering
\includegraphics[width=0.5\textwidth]{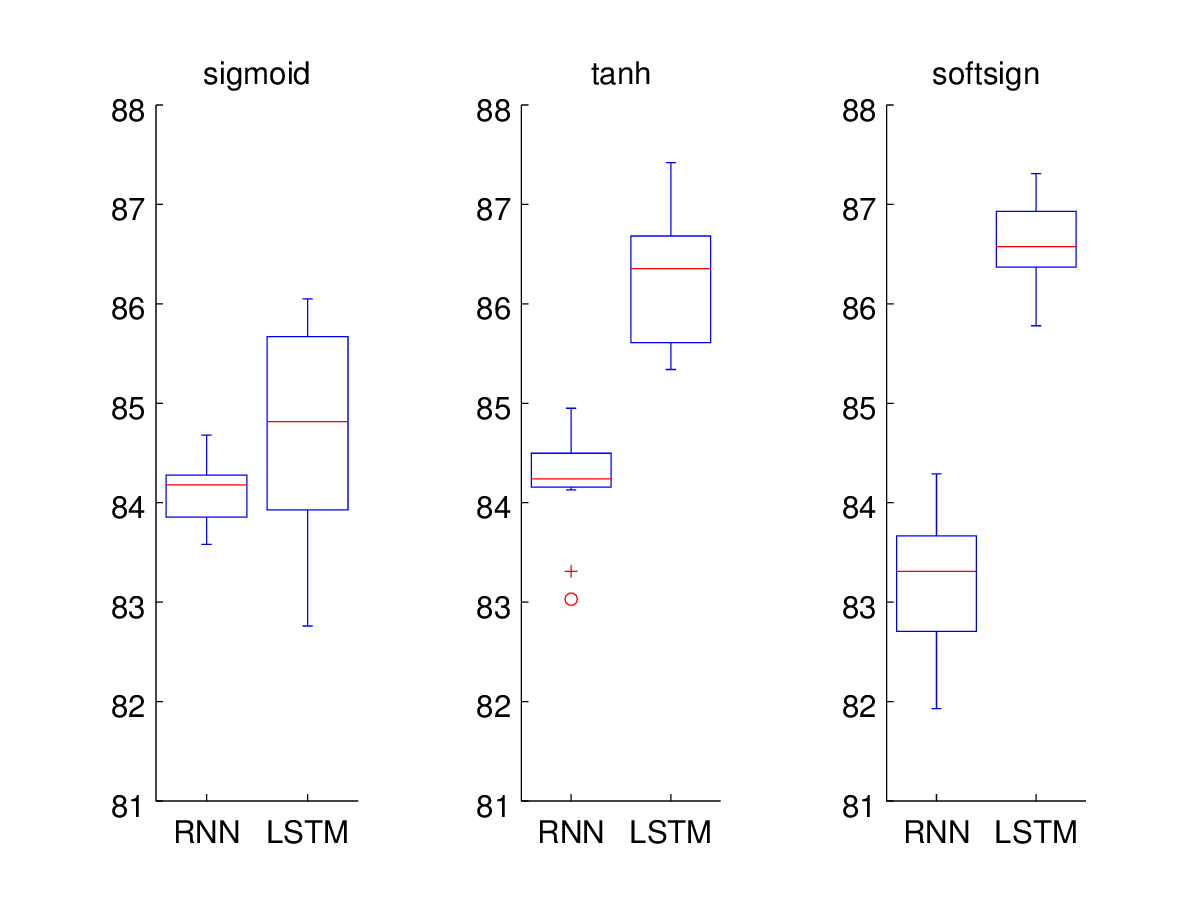}
\caption{Boxplot of accuracies of 10 runs of RNN and LSTM-RNN on the test set in the binary classification task.
(LSTM stands for LSTM-RNN.)}
\label{figure 2c LSTM vs RNN}
\end{figure}

\paragraph{Results}
Figure~\ref{figure 5c LSTM vs RNN} and ~\ref{figure 2c LSTM vs RNN}
show the statistics of the accuracies of the final networks on the test set in 
the fine-grained classification task and binary classification task, respectively.

It can be seen that LSTM-RNN outperformed RNN 
when using the tanh or softsign activation functions. With the sigmoid activation function, 
the difference is not so clear, but it seems that LSTM-RNN performed slightly better. 
Tanh-LSTM-RNN and softsign-LSTM-RNN  have the highest median accuracies 
(48.1 and 86.4) in the fine-grained classification task and in the binary 
classification task, respectively.

With the RNN model, it is surprising to see that the sigmoid function performed well, 
comparably with the other two functions in the fine-grained task, and even better than 
the softsign function in the binary task, given that it was not often chosen in recent work. 
The softsign function, which was shown to work better than tanh for 
deep networks \cite{glorot2010understanding}, however, did not yield improvements in this experiment. 

With the LSTM-RNN model, the tanh function, in general, worked best whereas 
the sigmoid function was the worst. This result agrees with the common choice for
this activation function for the LSTM architecture in  recurrent network research
\cite{gers2001long,sutskever2014sequence}.

\subsection{Compared against other Models}

We compare LSTM-RNN (using tanh) in the previous
experiment against existing models:
Naive Bayes with bag of bigram features (BiNB), Recursive neural tensor network (RNTN) 
\cite{socher2013recursive}, 
Convolutional neural network (CNN) \cite{kim:2014:EMNLP2014}, 
Dynamic convolutional neural network (DCNN) \cite{kalchbrenner-grefenstette-blunsom:2014:P14-1}, 
paragraph vectors (PV) \cite{le2014distributed}, 
and Deep RNN (DRNN) \cite{irsoy2014deep}. 

Among them, BiNB is the only one that is not a neural net model. 
RNTN and DRNN are two extensions of RNN. Whereas RNTN, which keeps the structure 
of the RNN, uses both matrix-vector multiplication and tensor product for the composition purpose, 
DRNN makes the net deeper by concatenating more than one RNNs horizontally. 
CNN, DCNN and PV do not rely on syntactic trees. 
CNN uses a convolutional layer and a max-pooling layer 
to handle sequences with different lengths. 
DCNN is hierarchical in the sense that it stacks more than one convolutional layers 
with k-max pooling layers in between. In PV, a sentence 
(or document) is represented as an input vector to predict which words appear in it.

Table~\ref{table compare} (above the dashed line) shows 
the accuracies of those models. 
The accuracies of LSTM-RNN was taken from the network
achieving the highest performance out of 10 runs on the 
development set. The accuracies of the other models are
copied from the corresponding papers. LSTM-RNN clearly performed 
worse than DCNN, PV, DRNN in both tasks, and worse than 
CNN in the binary task.

\begin{table}
\centering
\begin{tabular}{lcc}
 Model              &   Fine-grained    &   Binary \\ \hline
 
 BiNB               &   41.9            &   83.1 \\
 RNTN               &   45.7            &   85.4 \\
 CNN                &   48.0            &   \underline{88.1} \\
 DCNN               &   48.5            &   86.8 \\
 PV                 &   48.7            &   87.8 \\
 DRNN               &   \underline{49.8}   &   86.6 \\ \hline
 
 \multicolumn{3}{l}{\textit{with GloVe-100D}} \\
 LSTM-RNN &   48.0            &   86.2 \\ \hdashline
 
 \multicolumn{3}{l}{\textit{with GloVe-300D}} \\
 LSTM-RNN  &   \textbf{49.9}        &    \textbf{88.0}    \\
\end{tabular}
\caption{Accuracies of the (tanh) LSTM-RNN compared with other models.}
\label{table compare}
\end{table}

\subsection{Toward State-of-the-art with Better Word Embeddings}

We focus on DRNN, which is the most similar to LSTM-RNN among those four 
models CNN, DCNN, PV and DRNN. In fact, from the results reported in 
\newcite[Table 1a]{irsoy2014deep}, 
LSTM-RNN performed on par\footnote{
\newcite{irsoy2014deep} used the 300-D word2vec word embeddings trained on a 100B-word
corpus whereas we used the 100-D GloVe word embeddings trained on a 6B-word corpus. 
From the fact that they achieved the accuracy 46.1 with an RNN ($d=50$) in the 
fine-grained task and 85.3 in the binary task, 
and our implementation of RNN ($d=50$) performed worse (see Table~\ref{figure 5c LSTM vs RNN}
and \ref{figure 2c LSTM vs RNN}), we conclude that the 100-D GloVe word embeddings 
are not more suitable than the 300-D word2vec word embeddings.}
with their 1-layer-DRNN ($d=340$)
using dropout, which is to randomly remove some neurons during training. 
Dropout is a powerful technique to train neural networks, 
not only because it plays a role as a strong regularization method to prohibit 
neurons co-adapting, but it is also considered a 
technique to efficiently make an ensemble of a large number of shared weight neural networks
\cite{srivastava2014dropout}. Thanks to dropout, \newcite{irsoy2014deep} boosted 
the accuracy of a 3-layer-DRNN with $d=200$ from 46.06 to 49.5 in the fine-grained task.

In the second experiment, we tried to boost the accuracy of the LSTM-RNN model. 
Inspired by \newcite{irsoy2014deep}, we tried using dropout and 
better word embeddings. Dropout, however, did not work with LSTM. The reason 
might be that dropout corrupted its memory, thus making training more difficult. 
Better word embeddings did pay off, however. We used 300-D GloVe word embeddings trained on a 840B-word corpus. 
Testing on the development set, we chose the same values for the hyper-parameters as in 
the first experiment, except setting learning rate $0.01$. 
We also run the model 10 times and 
selected  the networks getting the highest accuracies on the development set. 
Table~\ref{table compare} (below the dashed line) shows the results. Using 
the 300-D GloVe word embeddings was very helpful: 
LSTM-RNN performed on par with DRNN 
in the fine-grained task, and with CNN in the binary task. Therefore, 
taking into account both tasks, LSTM-RNN with the 300-D GloVe word embeddings 
outperformed all other models.

\section{Discussion and Conclusion}
\label{section conclusion}

We proposed a new composition method for the recursive neural network (RNN) model 
by extending the long short-term memory (LSTM) architecture which is widely used 
in recurrent neural network research. 

The question is why LSTM-RNN performed better than the traditional RNN. Here, 
based on the fact that the LSTM for RNNs should work very similarly to LSTM for 
recurrent neural networks, we borrow the argument given in 
\newcite[Section 3.2]{bengio2013advances} to answer the question. Bengio 
explains that the LSTM behaves like low-pass filter 
``hence they can be used to focus certain units on different frequency regions of the data''. 
This suggests that the LSTM plays a role as a lossy compressor which is to 
keep global information by focusing on low frequency regions and remove noise 
by ignoring high frequency regions. So composition in this case could be seen as 
compression, like the recursive auto-encoder (RAE) \cite{socher_dynamic_2011}. Because 
pre-training an RNN as an RAE can boost the overall performance 
\cite{socher_dynamic_2011,socher2011semi}, seeing LSTM as a compressor  
might explain why the LSTM-RNN worked better than RNN without pre-training. 

Comparing LSTM-RNN against DRNN \cite{irsoy2014deep} gives us a hint about how to 
improve our model. From the experimental results, LSTM-RNN without the 
300-D GloVe word embeddings performed worse than DRNN, while DRNN gained a significant improvement 
thanks to dropout. 
Finding a method like dropout that does not  corrupt the LSTM memory 
might boost the overall performance significantly and will be a topic for our future work.

\section*{Acknowledgments}
We thank three anonymous reviewers for helpful comments.

\newpage

\bibliographystyle{naaclhlt2015}
\bibliography{ref}

\begin{thebibliography}{}

\bibitem[\protect\citename{Baroni \bgroup et al.\egroup
  }2013]{baroni_frege_2013}
Marco Baroni, Raffaella Bernardi, and Roberto Zamparelli.
\newblock 2013.
\newblock Frege in space: A program for compositional distributional semantics.
\newblock In A.~Zaenen, B.~Webber, and M.~Palmer, editors, {\em Linguistic
  Issues in Language Technologies}. CSLI Publications, Stanford, CA.

\bibitem[\protect\citename{Bengio \bgroup et al.\egroup
  }2013]{bengio2013advances}
Yoshua Bengio, Nicolas Boulanger-Lewandowski, and Razvan Pascanu.
\newblock 2013.
\newblock Advances in optimizing recurrent networks.
\newblock In {\em Acoustics, Speech and Signal Processing (ICASSP), 2013 IEEE
  International Conference on}, pages 8624--8628. IEEE.

\bibitem[\protect\citename{Cybenko}1989]{cybenko1989approximation}
George Cybenko.
\newblock 1989.
\newblock Approximation by superpositions of a sigmoidal function.
\newblock {\em Mathematics of control, signals and systems}, 2(4):303--314.

\bibitem[\protect\citename{Duchi \bgroup et al.\egroup
  }2011]{duchi2011adaptive}
John Duchi, Elad Hazan, and Yoram Singer.
\newblock 2011.
\newblock Adaptive subgradient methods for online learning and stochastic
  optimization.
\newblock {\em The Journal of Machine Learning Research}, pages 2121--2159.

\bibitem[\protect\citename{Elman}1990]{elman_finding_1990}
Jeffrey~L. Elman.
\newblock 1990.
\newblock Finding structure in time.
\newblock {\em Cognitive science}, 14(2):179--211.

\bibitem[\protect\citename{Gers}2001]{gers2001long}
Felix Gers.
\newblock 2001.
\newblock Long short-term memory in recurrent neural networks.
\newblock {\em Unpublished PhD dissertation, {\'E}cole Polytechnique
  F{\'e}d{\'e}rale de Lausanne, Lausanne, Switzerland}.

\bibitem[\protect\citename{Glorot and Bengio}2010]{glorot2010understanding}
Xavier Glorot and Yoshua Bengio.
\newblock 2010.
\newblock Understanding the difficulty of training deep feedforward neural
  networks.
\newblock In {\em International conference on artificial intelligence and
  statistics}, pages 249--256.

\bibitem[\protect\citename{Goller and K\"{u}chler}1996]{goller_learning_1996}
Christoph Goller and Andreas K\"{u}chler.
\newblock 1996.
\newblock Learning task-dependent distributed representations by
  backpropagation through structure.
\newblock In {\em International Conference on Neural Networks}, pages 347--352.
  IEEE.

\bibitem[\protect\citename{Graves}2012]{graves2012supervised}
Alex Graves.
\newblock 2012.
\newblock {\em Supervised sequence labelling with recurrent neural networks},
  volume 385.
\newblock Springer.

\bibitem[\protect\citename{Hochreiter and Schmidhuber}1997]{hochreiter1997long}
Sepp Hochreiter and J{\"u}rgen Schmidhuber.
\newblock 1997.
\newblock Long short-term memory.
\newblock {\em Neural computation}, 9(8):1735--1780.

\bibitem[\protect\citename{Hochreiter \bgroup et al.\egroup
  }2001]{Hochreiter2001}
S.~Hochreiter, Y.~Bengio, P.~Frasconi, and J.~Schmidhuber.
\newblock 2001.
\newblock Gradient flow in recurrent nets: the difficulty of learning long-term
  dependencies.
\newblock In Kremer and Kolen, editors, {\em A Field Guide to Dynamical
  Recurrent Neural Networks}. IEEE Press.

\bibitem[\protect\citename{Irsoy and Cardie}2014]{irsoy2014deep}
Ozan Irsoy and Claire Cardie.
\newblock 2014.
\newblock Deep recursive neural networks for compositionality in language.
\newblock In {\em Advances in Neural Information Processing Systems}, pages
  2096--2104.

\bibitem[\protect\citename{Kalchbrenner \bgroup et al.\egroup
  }2014]{kalchbrenner-grefenstette-blunsom:2014:P14-1}
Nal Kalchbrenner, Edward Grefenstette, and Phil Blunsom.
\newblock 2014.
\newblock A convolutional neural network for modelling sentences.
\newblock In {\em Proceedings of the 52nd Annual Meeting of the Association for
  Computational Linguistics (Volume 1: Long Papers)}, pages 655--665,
  Baltimore, Maryland, June. Association for Computational Linguistics.

\bibitem[\protect\citename{Kim}2014]{kim:2014:EMNLP2014}
Yoon Kim.
\newblock 2014.
\newblock Convolutional neural networks for sentence classification.
\newblock In {\em Proceedings of the 2014 Conference on Empirical Methods in
  Natural Language Processing (EMNLP)}, pages 1746--1751, Doha, Qatar, October.
  Association for Computational Linguistics.

\bibitem[\protect\citename{Le and Mikolov}2014]{le2014distributed}
Quoc Le and Tomas Mikolov.
\newblock 2014.
\newblock Distributed representations of sentences and documents.
\newblock In {\em Proceedings of the 31st International Conference on Machine
  Learning (ICML-14)}, pages 1188--1196.

\bibitem[\protect\citename{Le and Zuidema}2014a]{le2014the}
Phong Le and Willem Zuidema.
\newblock 2014a.
\newblock The inside-outside recursive neural network model for dependency
  parsing.
\newblock In {\em Proceedings of the 2014 Conference on Empirical Methods in
  Natural Language Processing}. Association for Computational Linguistics.

\bibitem[\protect\citename{Le and Zuidema}2014b]{leinside}
Phong Le and Willem Zuidema.
\newblock 2014b.
\newblock Inside-outside semantics: A framework for neural models of semantic
  composition.
\newblock In {\em NIPS 2014 Workshop on Deep Learning and Representation
  Learning}.

\bibitem[\protect\citename{Mikolov \bgroup et al.\egroup
  }2010]{mikolov2010recurrent}
Tomas Mikolov, Martin Karafi{\'a}t, Lukas Burget, Jan Cernock{\`y}, and Sanjeev
  Khudanpur.
\newblock 2010.
\newblock Recurrent neural network based language model.
\newblock In {\em INTERSPEECH}, pages 1045--1048.

\bibitem[\protect\citename{Mitchell and Lapata}2009]{mitchell2009language}
Jeff Mitchell and Mirella Lapata.
\newblock 2009.
\newblock Language models based on semantic composition.
\newblock In {\em Proceedings of the 2009 Conference on Empirical Methods in
  Natural Language Processing}, pages 430--439.

\bibitem[\protect\citename{Paulus \bgroup et al.\egroup
  }2014]{paulus2014global}
Romain Paulus, Richard Socher, and Christopher~D Manning.
\newblock 2014.
\newblock Global belief recursive neural networks.
\newblock In {\em Advances in Neural Information Processing Systems}, pages
  2888--2896.

\bibitem[\protect\citename{Pennington \bgroup et al.\egroup
  }2014]{pennington2014GloVe}
Jeffrey Pennington, Richard Socher, and Christopher~D Manning.
\newblock 2014.
\newblock Glove: Global vectors for word representation.
\newblock {\em Proceedings of the Empiricial Methods in Natural Language
  Processing (EMNLP 2014)}, 12.

\bibitem[\protect\citename{Rumelhart \bgroup et al.\egroup
  }1988]{rumelhart1988learning}
David~E Rumelhart, Geoffrey~E Hinton, and Ronald~J Williams.
\newblock 1988.
\newblock Learning representations by back-propagating errors.
\newblock {\em Cognitive modeling}, 5.

\bibitem[\protect\citename{Socher \bgroup et al.\egroup
  }2010]{socher_learning_2010}
Richard Socher, Christopher~D. Manning, and Andrew~Y. Ng.
\newblock 2010.
\newblock Learning continuous phrase representations and syntactic parsing with
  recursive neural networks.
\newblock In {\em Proceedings of the {NIPS-2010} Deep Learning and Unsupervised
  Feature Learning Workshop}.

\bibitem[\protect\citename{Socher \bgroup et al.\egroup
  }2011a]{socher_dynamic_2011}
Richard Socher, Eric~H. Huang, Jeffrey Pennington, Andrew~Y. Ng, and
  Christopher~D. Manning.
\newblock 2011a.
\newblock Dynamic pooling and unfolding recursive autoencoders for paraphrase
  detection.
\newblock {\em Advances in Neural Information Processing Systems}, 24:801--809.

\bibitem[\protect\citename{Socher \bgroup et al.\egroup
  }2011b]{socher_parsing_2011}
Richard Socher, Cliff~C. Lin, Andrew~Y. Ng, and Christopher~D. Manning.
\newblock 2011b.
\newblock Parsing natural scenes and natural language with recursive neural
  networks.
\newblock In {\em Proceedings of the 26th International Conference on Machine
  Learning}, volume~2.

\bibitem[\protect\citename{Socher \bgroup et al.\egroup }2011c]{socher2011semi}
Richard Socher, Jeffrey Pennington, Eric~H Huang, Andrew~Y Ng, and
  Christopher~D Manning.
\newblock 2011c.
\newblock Semi-supervised recursive autoencoders for predicting sentiment
  distributions.
\newblock In {\em Proceedings of the Conference on Empirical Methods in Natural
  Language Processing}, pages 151--161.

\bibitem[\protect\citename{Socher \bgroup et al.\egroup
  }2013a]{socher2013parsing}
Richard Socher, John Bauer, Christopher~D Manning, and Andrew~Y Ng.
\newblock 2013a.
\newblock Parsing with compositional vector grammars.
\newblock In {\em Proceedings of the 51st Annual Meeting of the Association for
  Computational Linguistics}, pages 455--465.

\bibitem[\protect\citename{Socher \bgroup et al.\egroup
  }2013b]{socher2013recursive}
Richard Socher, Alex Perelygin, Jean~Y Wu, Jason Chuang, Christopher~D Manning,
  Andrew~Y Ng, and Christopher Potts.
\newblock 2013b.
\newblock Recursive deep models for semantic compositionality over a sentiment
  treebank.
\newblock In {\em Proceedings EMNLP}.

\bibitem[\protect\citename{Srivastava \bgroup et al.\egroup
  }2014]{srivastava2014dropout}
Nitish Srivastava, Geoffrey Hinton, Alex Krizhevsky, Ilya Sutskever, and Ruslan
  Salakhutdinov.
\newblock 2014.
\newblock Dropout: A simple way to prevent neural networks from overfitting.
\newblock {\em The Journal of Machine Learning Research}, 15(1):1929--1958.

\bibitem[\protect\citename{Sutskever \bgroup et al.\egroup
  }2014]{sutskever2014sequence}
Ilya Sutskever, Oriol Vinyals, and Quoc~VV Le.
\newblock 2014.
\newblock Sequence to sequence learning with neural networks.
\newblock In {\em Advances in Neural Information Processing Systems}, pages
  3104--3112.

\bibitem[\protect\citename{Tai \bgroup et al.\egroup }2015]{tai2015improved}
Kai~Sheng Tai, Richard Socher, and Christopher~D Manning.
\newblock 2015.
\newblock Improved semantic representations from tree-structured long
  short-term memory networks.
\newblock {\em arXiv preprint arXiv:1503.00075}.

\bibitem[\protect\citename{Werbos}1990]{werbos1990backpropagation}
Paul~J Werbos.
\newblock 1990.
\newblock Backpropagation through time: what it does and how to do it.
\newblock {\em Proceedings of the IEEE}, 78(10):1550--1560.

\bibitem[\protect\citename{Zhu \bgroup et al.\egroup }2015]{zhu2015long}
Xiaodan Zhu, Parinaz Sobhani, and Hongyu Guo.
\newblock 2015.
\newblock Long short-term memory over tree structures.
\newblock {\em arXiv preprint arXiv:1503.04881}.

\end{thebibliography}

\end{document}